\documentclass[10pt,twocolumn,letterpaper]{article}

\usepackage{cvpr}
\usepackage{times}
\usepackage{epsfig}
\usepackage{graphicx}
\usepackage{amsmath}
\usepackage{amssymb}
\usepackage{multirow}

\usepackage[pagebackref=true,breaklinks=true,letterpaper=true,colorlinks,bookmarks=false]{hyperref}

\cvprfinalcopy 

\def\cvprPaperID{15} 

\ifcvprfinal\fi
\begin{document}
\title{\vspace{-0.95cm}Autonomous Human Activity Classification from Ego-vision Camera and Accelerometer Data \vspace{-0.3cm}}

\author{Yantao Lu\\
Dept. of Electrical Eng. and Computer Science\\
Syracuse University\\
{\tt\small ylu25@syr.edu} \vspace{-0.45cm}
\and
Senem Velipasalar\\
Dept. of Electrical Eng. and Computer Science\\
Syracuse University\\
{\tt\small svelipas@syr.edu} \vspace{-0.45cm}
}

\maketitle

\begin{abstract} \vspace{-0.2cm}
  There has been significant amount of research work on human activity classification relying either on Inertial Measurement Unit (IMU) data or data from static cameras providing a third-person view. Using only IMU data limits the variety and complexity of the activities that can be detected. For instance, the sitting activity can be detected by IMU data, but it cannot be determined whether the subject has sat on a chair or a sofa, or where the subject is. To perform fine-grained activity classification from egocentric videos, and to distinguish between activities that cannot be differentiated by only IMU data, we present an autonomous and robust method using data from both ego-vision cameras and IMUs. In contrast to convolutional neural network-based approaches, we propose to employ capsule networks to obtain features from egocentric video data. Moreover, Convolutional Long Short Term Memory framework is employed both on egocentric videos and IMU data to capture temporal aspect of actions. We also propose a genetic algorithm-based approach to autonomously and systematically set various network parameters, rather than using manual settings. Experiments have been performed to perform 9- and 26-label activity classification, and the proposed method, using autonomously set network parameters, has provided very promising results, achieving overall accuracies of 86.6\% and 77.2\%, respectively. The proposed approach combining both modalities also provides increased accuracy compared to using only egovision data and only IMU data.
\end{abstract}

\vspace{-0.4cm}
\section{Proposed Method} \label{sec:proposedMethod} 
\begin{figure}[h!]
\vspace{-0.2cm}
  \centering
   \centerline{\includegraphics[width=.38\textwidth]{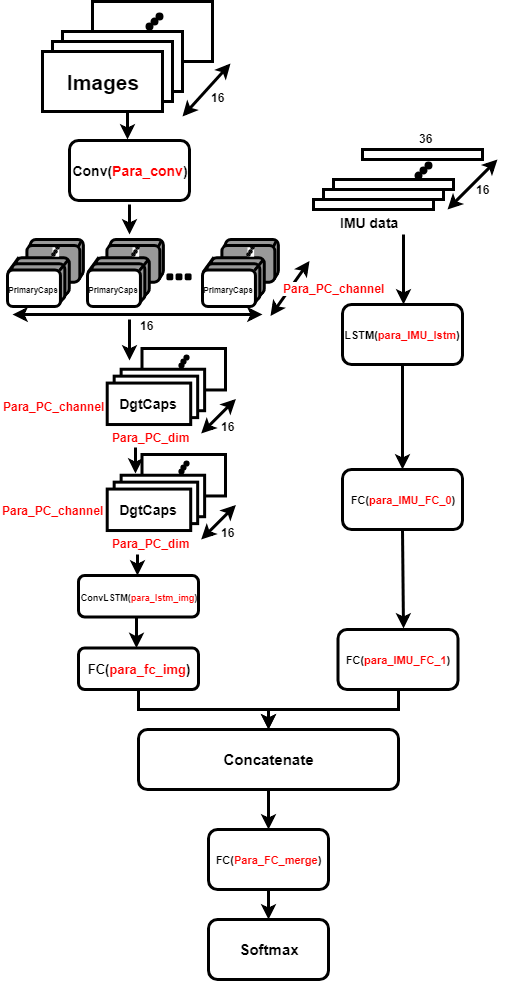}}
\caption{{\small{Details of the proposed architecture.}}}
\label{fig:network_details}
\vspace{-0.2cm}
\end{figure}

\vspace{-0.15cm}
We present a new model architecture to process first-person, also known as egocentric, images and IMU data. The proposed architecture can be seen in Fig.~\ref{fig:network_details}. It is composed of our proposed recurrent CapsNet (for processing images), an LSTM network (for processing IMU data), and fully connected layers. In addition, we also propose and apply a Genetic Algorithm (GA)-based approach to autonomously and simultaneously optimize multiple parameters of our network architecture. These parameters are shown in parentheses with red color in Fig.~\ref{fig:network_details}. For instance, the parameters for the fully connected layers, and the primary capsules are examples of the parameters autonomously set by our proposed approach.

Sabour et al.~\cite{Sabour2017_CapsNet} introduced the Capsule Networks (CapsNets) to explore spatial relationships between features, and reported state-of-the-art performance on the MNIST database. CapsNets \cite{Sabour2017_CapsNet} were used for image classification on individual images, whereas our goal is to perform fine-grained activity classification from video data. Thus, in this paper, instead of using a single image with the original CapsNet, we propose a Recurrent CapsNet (RecCapsNet), which takes a sequence of images as input. We implement a 2D Convolutional LSTM (convLSTM)~\cite{Shi2017_convlstm} layer to extract features and capture the temporal aspect. For robustness, we use multiple digit/class layers instead of using only a single digit layer as was done in~\cite{Sabour2017_CapsNet}. In order to prevent gradient vanishing, we remove the squash function for digit layers and implement ReLu activation function instead.

As seen in Fig.~\ref{fig:network_details}, 16 consecutive images are passed through a 2D convolutional layer separately. The size of each input image is 36$\times$36. Then, the output for each image is sent to multiple primary capsules, the number of which is determined by our GA. When 16 consecutive images are formed, 50\% overlap is used throughout the video. The number of convolutional units for each primary capsule is also determined by the GA. The output from the primary capsule layer is then sent through two digit/class layers, whose parameters are set by the GA. We then apply a Convolutional LSTM layer, followed by a fully connected (FC) layer, for the analysis of the egocentric video data.

For the decoder part, we apply 16 sub-decoders to each image frame. Each sub-decoder has the same structure with the decoder of the original CapsNet except the sigmoid output is 1296 ($36\times 36$).

As for the IMU, data from 16 consecutive time frames is used. Each of the 16 IMU data vectors has 36 components obtained by concatenating data from the four IMU sensors. Each IMU sensor contributes nine entries from accelerometer, gyroscope and magnetometer measurements. The time stamps are provided for camera and IMU data in the CMU-MMAC dataset. To align the camera and IMU data, for a given camera image, the IMU time stamp that is closest to the camera time stamp is found.

\vspace{-0.3cm}
\subsection{Autonomously and Simultaneously Refining the Network Parameters} \label{ssec:prmChoice} \vspace{-0.15cm}

\begin{figure}[h!]
\vspace{-0.2cm}
  \centering
   \centerline{\includegraphics[width=.42\textwidth]{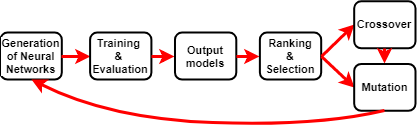}}
\caption{{\small{The structure of the proposed Genetic Algorithm}}}
\label{fig:GA}
\vspace{-0.3cm}
\end{figure}

The overall structure of the proposed method to refine the network parameters is shown in Fig.~\ref{fig:GA}. In this approach, a Genetic Algorithm (GA) is used to make a decision from a set of discrete choices. The complete set of network parameters refined by the GA together with the discrete set of values that they can take are shown in Table \ref{table:GA parameters1}.

\begin{table}[hb!]
\vspace{-0.3cm}
\caption{Parameters Autonomously Chosen by the GA}
\resizebox{\linewidth}{!}{
\begin{tabular}{l|l}
\hline
optimizers                 & \{``adam", ``rmsprop", ``adagrad", ``adadelta"\} \\
activation functions       & \{``relu", ``leaky relu", ``sigmoid", ``tanh"\}  \\
batch normalization        & \{True, False\}                              \\
dropout                    & \{True, False\}                              \\
max pooling                & \{True, False\}                              \\
kernel size                & \{3, 6, 9\}                                  \\
kernel stride              & \{1, 2, 3\}                                  \\
number of conv filters     & \{32, 64, 128 ... 512\}                      \\
number of dense neurons    & \{32, 64, 128, 256\}                         \\
number of lstm units       & \{16, 32, 64 ... 256\}                       \\
dimension of capsules      & \{2, 4, 8, 16\}                              \\
number of primary channels & \{16, 32, 64\}                               \\
number of conv layers      & \{3,6\}                            \\
number of dense layers     & \{1,3\}                            \\
number of LSTM layers      & \{1,3\}                            \\ \hline
\end{tabular}
}
\label{table:GA parameters1}
\vspace{-0.3cm}
\end{table}

\vspace{-0.4cm}
\subsubsection{Initial Population and Evaluation} \vspace{-0.2cm}
The first generation of the networks, $N^1 = \{N_1, N_2, ..., N_{n_m}\}$, where $n_m$ is the number of models, is generated by randomly choosing the values of parameters from the possible choices. The value of $n_m$ was set to be 10 in our experiments. Each generated network model $N_i$ is evaluated by the fitness function $f(N_i)$, which is a measure of the accuracy of each model.
\vspace{-0.4cm}
\subsubsection{Selection} \vspace{-0.2cm}
$t$-many top-ranked models are selected first, and then $r$-many models are selected randomly from the rest of the network models. Then, $d$-many models are dropped to prevent over-fitting and getting stuck at a local optimum. The remaining selected models are the parent models ($P$), which will be used to create new models for the next generation.
\vspace{-0.4cm}
\subsubsection{Crossover and Mutation} \vspace{-0.2cm}
Crossover is applied to generate $n_m$-many child network models from the parents. As opposed to always choosing two parents randomly from the parent pool, we associate a counter $C_P$ with each parent $P$, and initialize it to zero. This counter is incremented by one each time a parent is used for crossover. First, two parents are selected randomly from the $t+r-d$ many parents. A new `child' network is generated from the parents via crossover, and the counters of the parents are incremented by one. Then, two parents, whose counter is still zero, are selected randomly from the parent pool. Another network is generated from them via crossover, and the counters of the parents are incremented. If there is only one network model left with counter equal to zero, and the number of children is still less than $n_m$, then this model is chosen as one of the parents, and the other parent is chosen randomly from the rest of the models who have a counter value of one. If there are no more parents left with counter equal to zero, and the number of children is still less than $n_m$, then two parents, whose counter is one, are picked randomly, and their counter is incremented to two after crossover. This process is repeated until the number of children models reaches $n_m$.

\begin{figure}[h!]
\vspace{-0.3cm}
  \centering
   \centerline{\includegraphics[width=.22\textwidth]{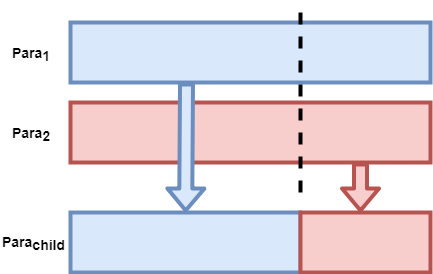}}
\caption{{\small{Crossover process for the GA}}}
\label{fig:crossover}
\vspace{-0.35cm}
\end{figure}

The crossover between parent models $a$ and $b$ is performed, as illustrated in Fig.~\ref{fig:crossover}, by using a single-point crossover. After all the $n_m$-many child networks are obtained, mutation is performed. The values of the parameters corresponding to randomly chosen $k$-many indices are randomly changed to one of the possible choices shown in Table \ref{table:GA parameters1}. The value of $k$ was chosen to be $3$ in our experiments.

\vspace{-0.4cm}
\section{Experimental Results} \label{sec:exp} \vspace{-0.15cm}
We have used CMU Multi-Modal Activity (CMU-MMAC) database~\cite{Fernando2009_MMAC}, which contains data from multi-modal sensors monitoring human subjects preparing food. 25 subjects were recorded cooking five different recipes. The sensor modalities used for data collection include three high resolution static cameras, two low-resolution static cameras, one wearable camera, five microphones and IMUs. In our experiments, we used the egocentric camera data and the wired IMU data. We resized the image frames from camera to $36\times36$ pixels. We down-sampled the IMU data to make the measuring frequency the same with the egocentric camera (30 Hz). Then, we synchronized/aligned the IMU data with camera data.

\begin{figure}[hbt!]
\vspace{-0.1cm}
\centering
\includegraphics[width=0.8\linewidth]{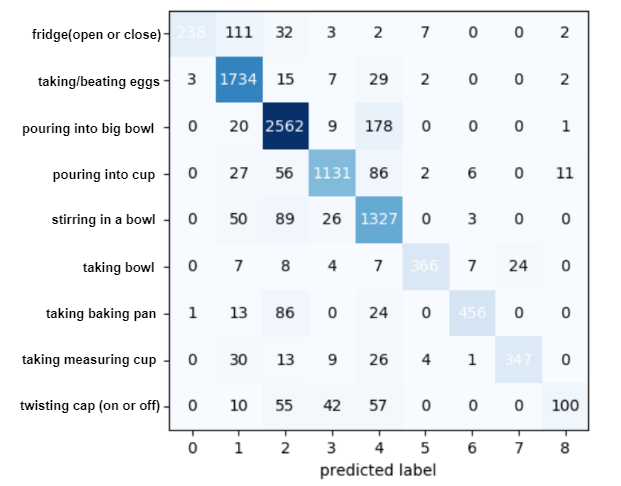}
\vspace{-0.1cm}
\caption{{\small{Confusion matrix for 9-class scenario.}}}
\label{fig:CM9}
\vspace{-0.25cm}
\end{figure}

\begin{figure}[hbt!]
\vspace{-0.1cm}
\centering
\includegraphics[width=0.95\linewidth]{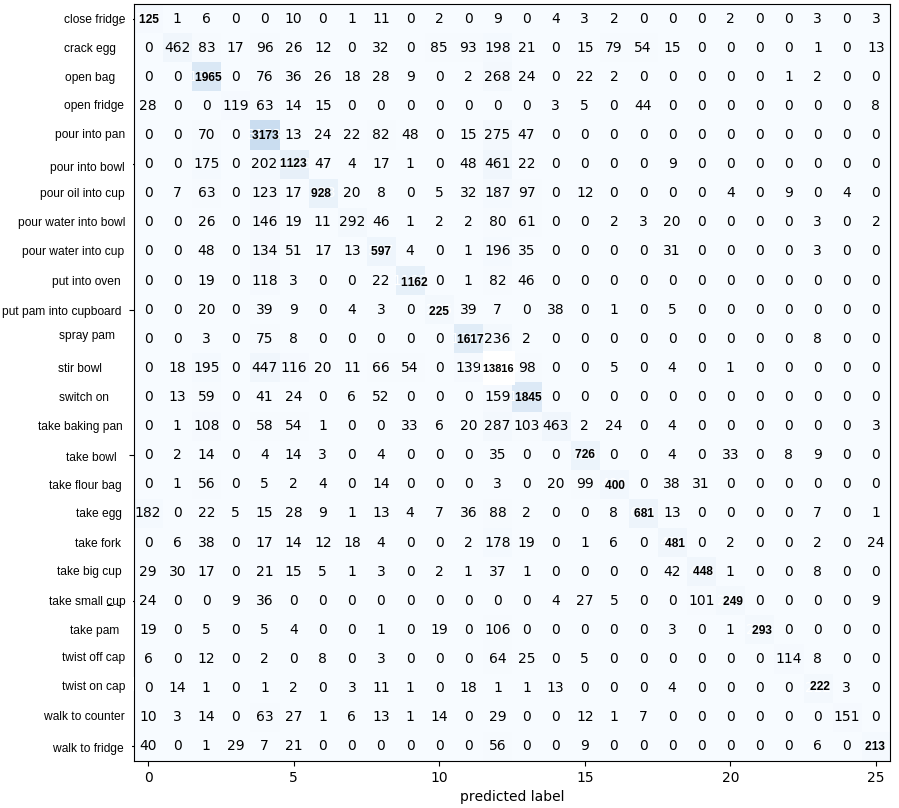}
\vspace{-0.2cm}
\caption{{\small{Confusion matrix for 26-class scenario.}}}
\label{fig:CM26}
\vspace{-0.55cm}
\end{figure}

\begin{figure}[tb!]
\begin{minipage}[b]{.02\columnwidth}
  \centering
  \centerline{\small{(1)}} \smallskip
\end{minipage}
\hfill
\begin{minipage}[b]{.31\columnwidth}
  \centering
  \centerline{\includegraphics[width=1\linewidth]{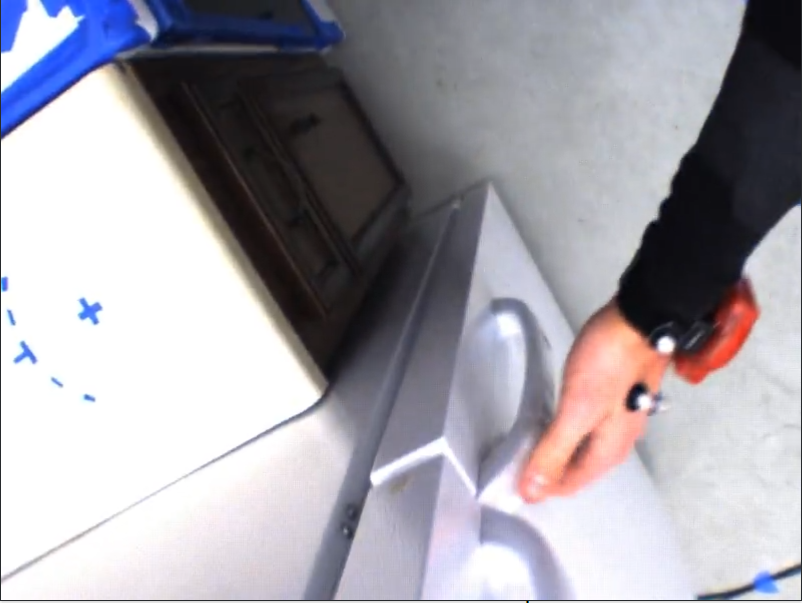}}
  \smallskip
\end{minipage}
\hfill
\begin{minipage}[b]{.31\columnwidth}
  \centering
  \centerline{\includegraphics[width=1\linewidth]{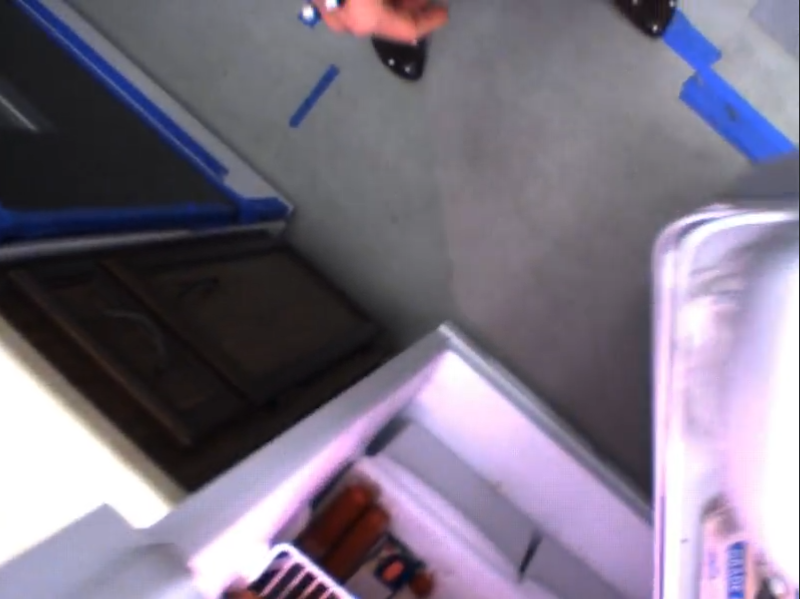}}
  \smallskip
\end{minipage}
\hfill
\begin{minipage}[b]{.31\columnwidth}
  \centering
  \centerline{\includegraphics[width=1\linewidth]{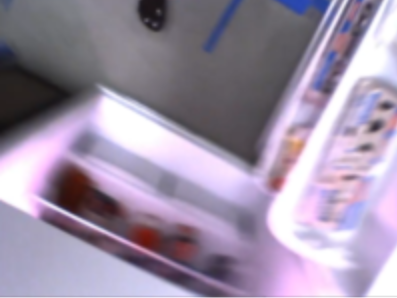}}
  \smallskip
\end{minipage} \\
\begin{minipage}[b]{.02\columnwidth}
  \centering
  \centerline{\small{(2)}} \smallskip
\end{minipage}
\hfill
\begin{minipage}[b]{.31\columnwidth}
  \centering
  \centerline{\includegraphics[width=1\linewidth]{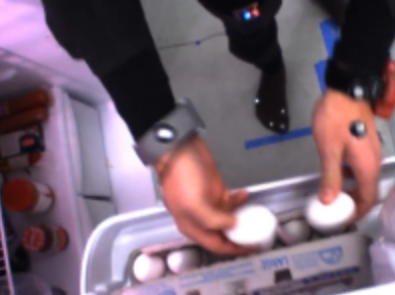}}
  \smallskip
\end{minipage}
\hfill
\begin{minipage}[b]{.31\columnwidth}
  \centering
  \centerline{\includegraphics[width=1\linewidth]{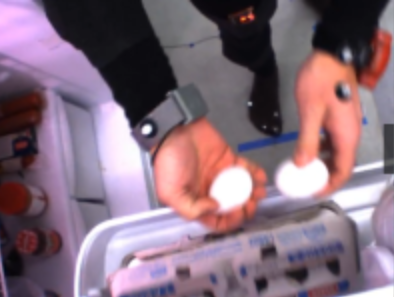}}
  \smallskip
\end{minipage}
\hfill
\begin{minipage}[b]{.31\columnwidth}
  \centering
  \centerline{\includegraphics[width=1\linewidth]{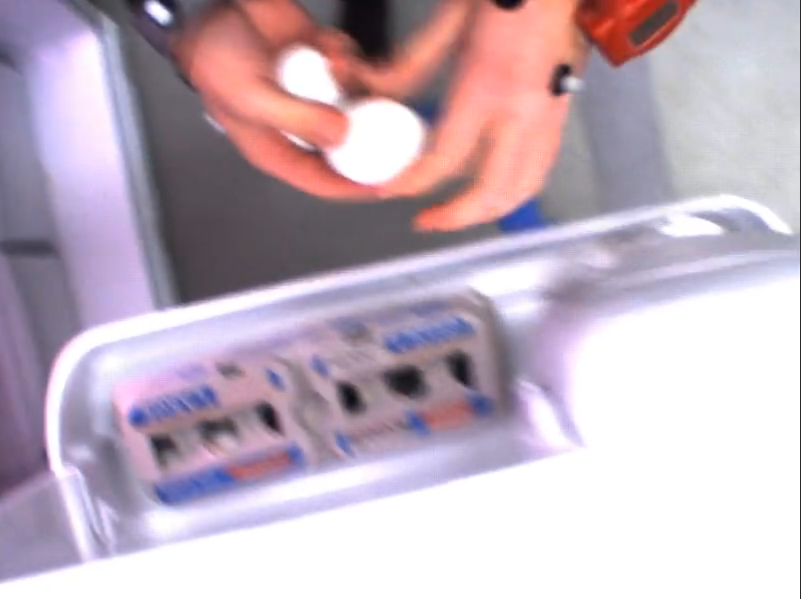}}
  \smallskip
\end{minipage} \\
\begin{minipage}[b]{.02\columnwidth}
  \centering
  \centerline{\small{(3)}} \smallskip
\end{minipage}
\hfill
\begin{minipage}[b]{.31\columnwidth}
  \centering
  \centerline{\includegraphics[width=1\linewidth]{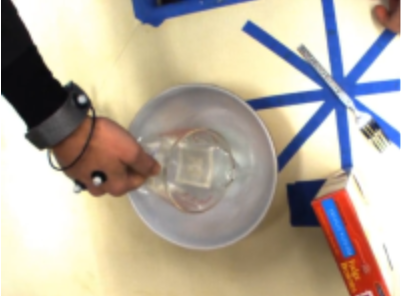}}
  \smallskip
\end{minipage}
\hfill
\begin{minipage}[b]{.31\columnwidth}
  \centering
  \centerline{\includegraphics[width=1\linewidth]{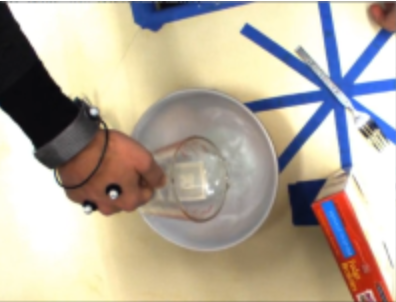}}
  \smallskip
\end{minipage}
\hfill
\begin{minipage}[b]{.31\columnwidth}
  \centering
  \centerline{\includegraphics[width=1\linewidth]{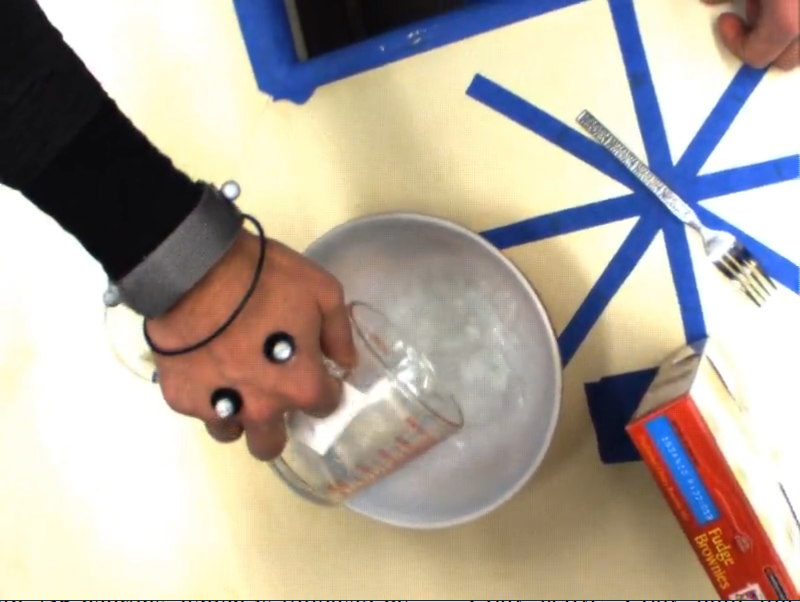}}
  \smallskip
\end{minipage}\\
\begin{minipage}[b]{.02\columnwidth}
  \centering
  \centerline{\small{(4)}} \smallskip
\end{minipage}
\hfill
\begin{minipage}[b]{.31\columnwidth}
  \centering
  \centerline{\includegraphics[width=1\linewidth]{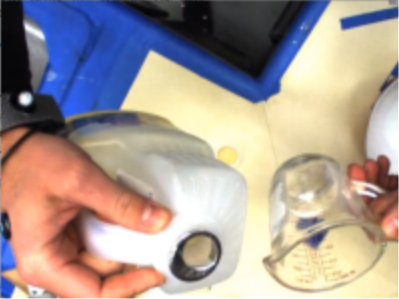}}
  \smallskip
\end{minipage}
\hfill
\begin{minipage}[b]{.31\columnwidth}
  \centering
  \centerline{\includegraphics[width=1\linewidth]{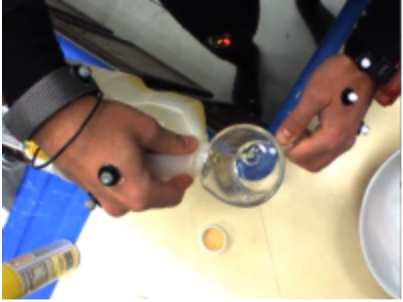}}
  \smallskip
\end{minipage}
\hfill
\begin{minipage}[b]{.31\columnwidth}
  \centering
  \centerline{\includegraphics[width=1\linewidth]{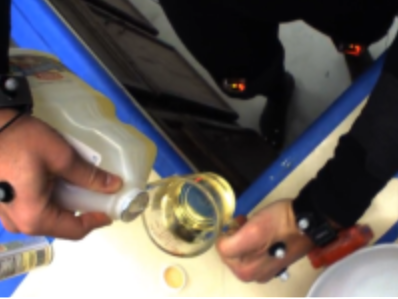}}
  \smallskip
\end{minipage}\\
\begin{minipage}[b]{.02\columnwidth}
  \centering
  \centerline{\small{(5)}} \smallskip
\end{minipage}
\hfill
\begin{minipage}[b]{.31\columnwidth}
  \centering
  \centerline{\includegraphics[width=1\linewidth]{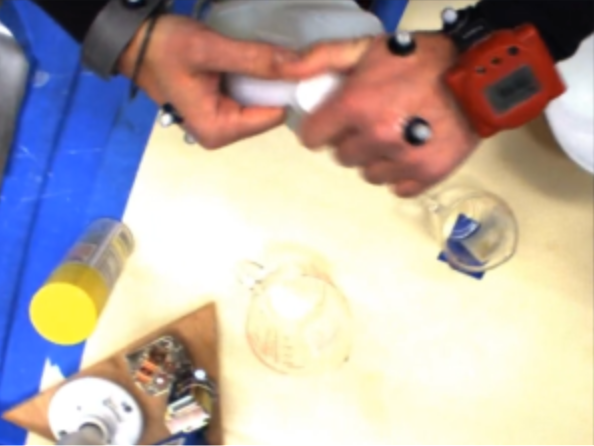}}
  \smallskip
\end{minipage}
\hfill
\begin{minipage}[b]{.31\columnwidth}
  \centering
  \centerline{\includegraphics[width=1\linewidth]{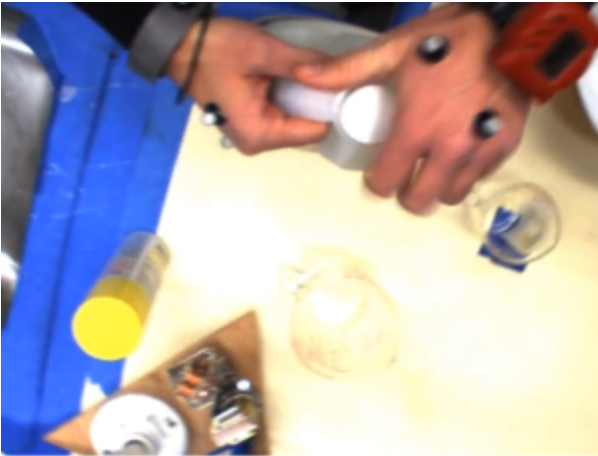}}
  \smallskip
\end{minipage}
\hfill
\begin{minipage}[b]{.31\columnwidth}
  \centering
  \centerline{\includegraphics[width=1\linewidth]{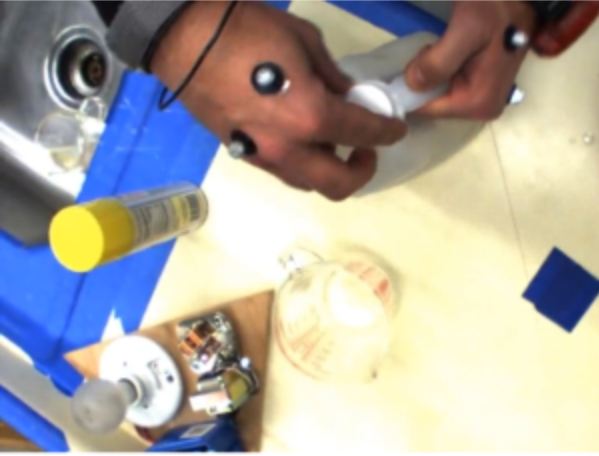}}
  \smallskip
\end{minipage} \\
\vspace{-0.4cm}
\caption{{\small{Example images from the CMU-MMAC dataset. Rows: (1) using fridge, (2) taking eggs, (3) pouring into big bowl, (4) pouring into a measuring cup, (5) twisting cap (on or off).}}}
\label{fig:expImages}
\vspace{-0.35cm}
\end{figure}

We performed two sets of experiments by using 9 and 26 different activity classes. The name of the activities for each case can be seen in the confusion matrices in Figures \ref{fig:CM9} and \ref{fig:CM26}. Example images are shown in Fig.~\ref{fig:expImages}. As can be seen, especially for the 26-class case, the activities involved are very close in the `activity space', and this fine-grain classification is a very challenging problem.

A total of 10 videos from five subjects (2 videos per subject) have been used for training and testing. Videos from each subject were randomly divided so that 70\%, 20\%, 10\% of the samples were allocated for training, validation and testing, respectively.


We first performed classification with manually preset network parameters, and then with the parameters determined autonomously by our GA-based approach described above.~The overall accuracies from these experiments are summarized in Table \ref{table:Accuracy}, wherein the accuracy is the ratio of all correctly classified instances to the total number of instances. When we use our proposed GA-based approach to autonomously set the various parameters of the network, this provides higher accuracy for both 9-class and 26-class labeling. Thus, the remainder of the results are presented for when the parameters are set with our GA-based approach.
\begin{table}[hb!]
\vspace{-0.25cm}
\centering
\caption{Overall accuracies for the 9- and 26-class labeling with and without using the proposed GA-based parameter setting}
\resizebox{\linewidth}{!}{
\begin{tabular}{|l|cc|cc|}
\hline
         & \multicolumn{2}{c|}{9-class} & \multicolumn{2}{c|}{26-class} \\
         & Preset prm.        & GA-based prm.         & Preset prm.        & GA-based prm.           \\ \hline
Acc. & 84.2\%       & \textbf{86.6\%}       & 75.7\%        & \textbf{77.2\%}       \\ \hline
\end{tabular}
}
\label{table:Accuracy}
\vspace{-0.3cm}
\end{table}

The confusion matrices for the 9- and 26-class activity classification are shown in Fig.~\ref{fig:CM9} and \ref{fig:CM26}, respectively. When subjects interact with larger objects, and movements are faster, a higher accuracy is achieved compared to slower movements and interacting with smaller objects. For instance, it is harder to detect `twisting cap on' and `twisting cap off' actions, since the cap is always occluded by hand. As another example, actions such as cracking egg are harder to classify, since the egg is much smaller than the bowl.

\begin{figure}[tb!]
\begin{minipage}[b]{.32\columnwidth}
  \centering
  \centerline{\includegraphics[width=1\linewidth]{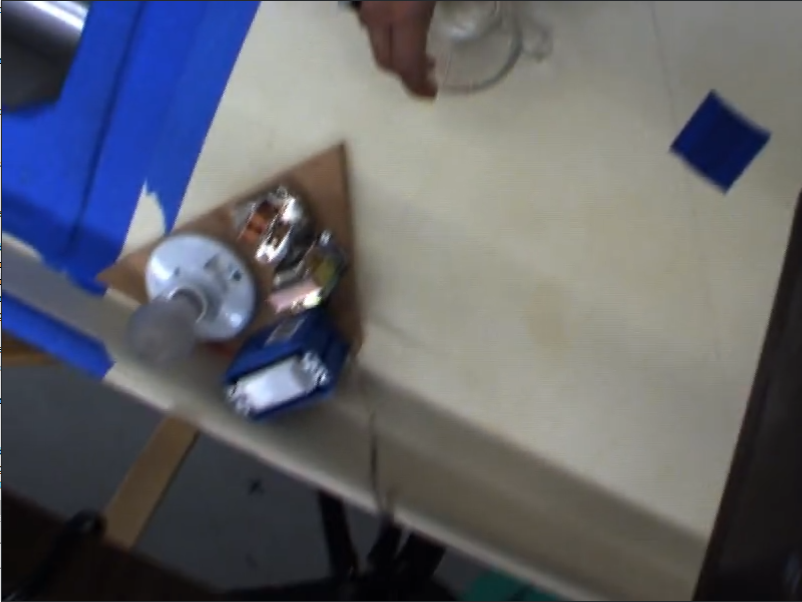}}
  \smallskip
\end{minipage}
\hfill
\begin{minipage}[b]{.32\columnwidth}
  \centering
  \centerline{\includegraphics[width=1\linewidth]{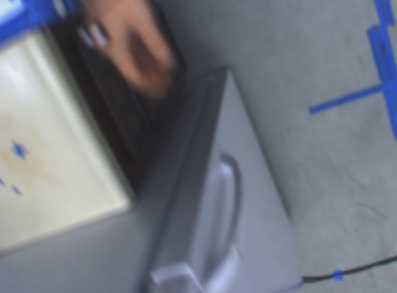}}
  \smallskip
\end{minipage}
\hfill
\begin{minipage}[b]{.32\columnwidth}
  \centering
  \centerline{\includegraphics[width=1\linewidth]{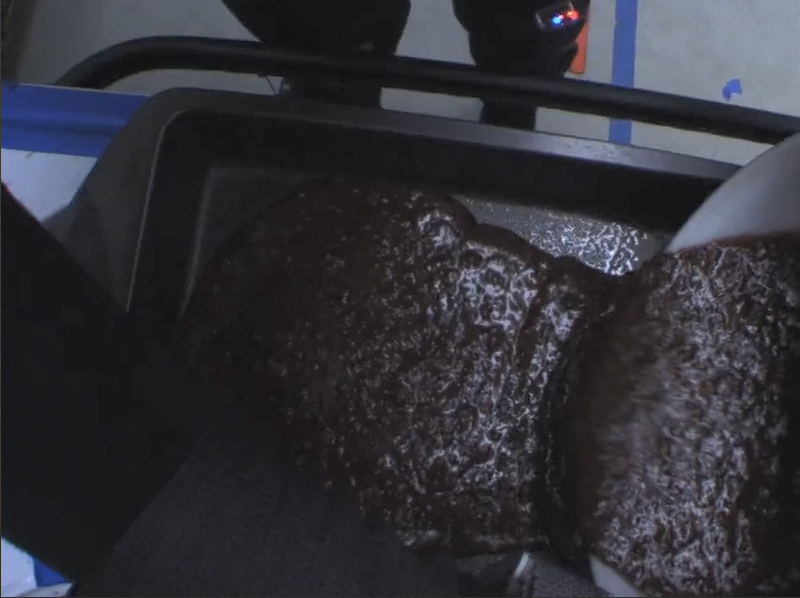}}
  \smallskip
\end{minipage}\\
\begin{minipage}[b]{.32\columnwidth}
  \centering
  \centerline{\includegraphics[width=1\linewidth]{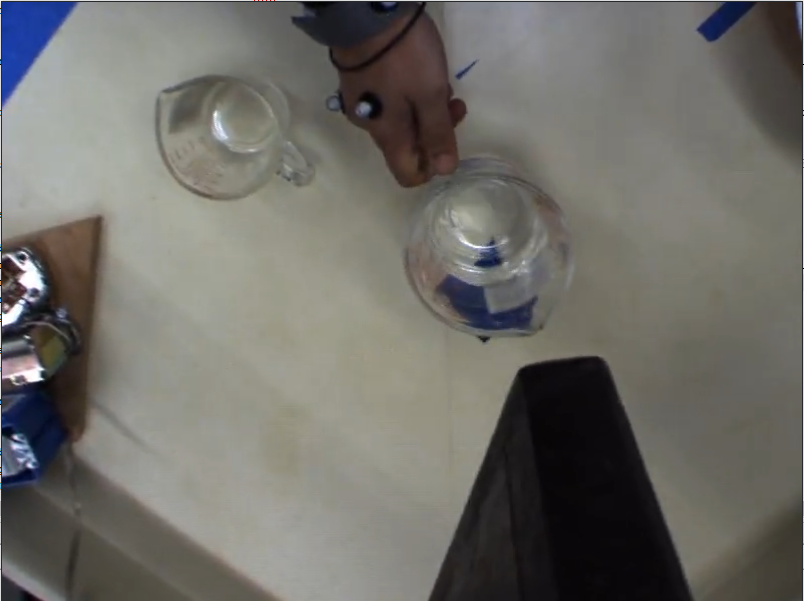}}
  \smallskip
\end{minipage}
\hfill
\begin{minipage}[b]{.32\columnwidth}
  \centering
  \centerline{\includegraphics[width=1\linewidth]{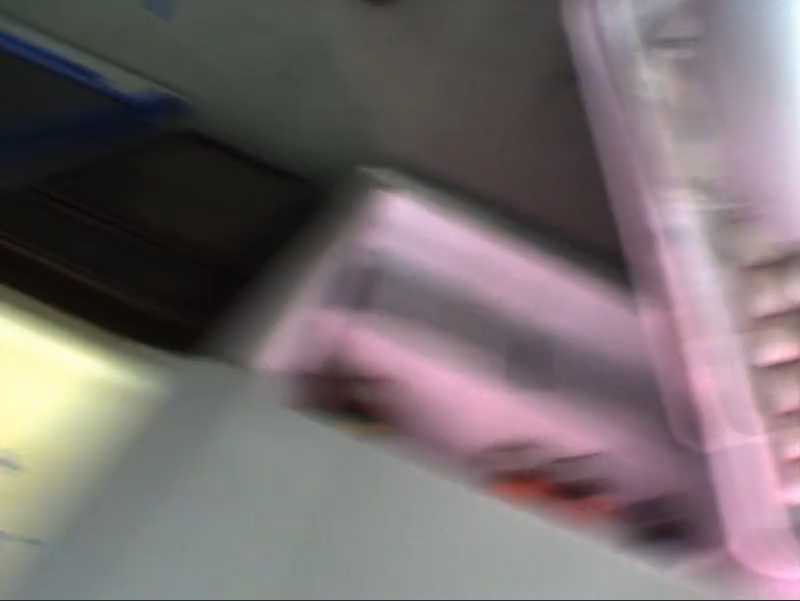}}
  \smallskip
\end{minipage}
\hfill
\begin{minipage}[b]{.32\columnwidth}
  \centering
  \centerline{\includegraphics[width=1\linewidth]{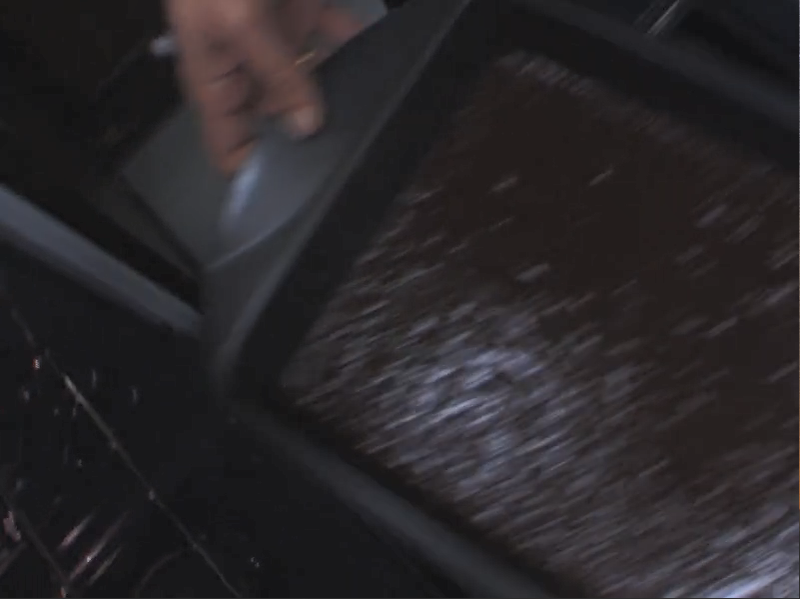}}
  \smallskip
\end{minipage}
\hfill \\
\vspace{-0.5cm}
\caption{{\small{Example challenging cases causing confusion. Columns: (1) taking a small cup (top) vs. big cup (bottom), (2) walking to fridge (top) vs. closing fridge (bottom), (3) pouring into pan (top) vs. putting pan into oven (bottom).}}}
\label{fig:diffImages}
\vspace{-0.65cm}
\end{figure}

In addition, as expected, higher overall precision and recall rates are achieved for 9-class labeling, since activities are much closer to each other and harder to differentiate for the 26-class labeling case. In Fig.~\ref{fig:diffImages}, we show example images for the activities that are confused with each other in the 26-class labeling case (based on the confusion matrix in Fig.~\ref{fig:CM26}). These images illustrate once more the difficulty of performing very fine-grained activity classification. The first row shows taking a small cup vs. big cup. The second row shows walking to the fridge vs. closing the fridge, and the third row shows pouring into pan vs. putting the pan into the oven. As can be seen, these are very similar looking activities, and the proposed approach still provides very promising results for the 26-class labeling.

After setting the various network parameters by our GA-based approach, we performed a comparison of our proposed Rec-CapsNets method with using VGG16 features. For this comparison, instead of employing the proposed RecCapsNet, we extracted image features from 16 consecutive image frames by using the convolutional layers of the CNN-based VGG16~\cite{VGG16} without the top layers. We also used CapsNet on individual frames. The results are summarized in Table \ref{table:comp9class} for 9-label classification. As can be seen, using our proposed RecCapsNet provides a higher accuracy then using the VGG16 features. Moreover, to show the improvement provided by using multiple sensor modalities, we also obtained results by using each sensor modality by itself, namely by using only IMU data and only camera data. As can be seen in Table \ref{table:comp9class}, the proposed approach provides 29.07\%, 20.29\% and 19.16\% increase in accuracy compared to using only IMU data, only egocentric camera data with VGG16 features and only egocentric camera data with CapsNet features, respectively.

\begin{table}[hb!] 
\vspace{-0.25cm}
\caption{Accuracy rates from different modalities and approaches for 9-label classification}
\resizebox{\linewidth}{!}{\begin{tabular}{c|c|c}
\hline
Sensor Modality & Method & Accuracy \\ \hline \hline
IMU only & LSTM & 57.57\% \\ \hline
\multirow{2}{*}{Camera only} & VGG16 & 66.35\%\\
& CapsNet & 67.48\%  \\ \hline
\multirow{2}{*}{Camera and IMU} & VGG16 \& LSTM & 82.97\%\\
& \textbf{RecCapsNet \& LSTM (Proposed)} & \textbf{86.64\%}  \\ \hline
\end{tabular}}
\label{table:comp9class}
\vspace{-0.35cm}
\end{table}

We also compared our results with two other works \cite{Spriggs2009Temporal}\cite{Soran15}, which use the same MMAC dataset. In general, a direct comparison would not be commensurate, since they either employ different sets of sensors and handcrafted features, or the annotations are different. The reported accuracy in \cite{Soran15}, from egocentric camera data, is 37.92\% for 28 classes. When data from wearable camera as well as the multiple static cameras, watching the subjects, are used, the reported average accuracy is 54.62\%~\cite{Soran15}. Spriggs et al.~\cite{Spriggs2009Temporal} report an accuracy of 57.8\% from 29 classes. Overall, our proposed method provides a significant improvement without relying on the static cameras watching the targets, which could also be important to alleviate privacy concerns. Moreover, using the proposed GA-based approach not only provides a way to systematically set the network parameters, but also improves the performance further compared to using the manually set parameters.

\vspace{-0.25cm}
\section{Conclusion} \vspace{-0.2cm}
We have presented an autonomous method for fine-grain activity classification by using data from both egovision cameras and IMUs. We have employed capsule networks to obtain features from egocentric videos.
We have also proposed a genetic algorithm-based approach to autonomously set various network parameters. \vspace{-0.15cm}

{\footnotesize
\bibliographystyle{ieee_fullname}
\bibliography{refs}

\begin{thebibliography}{1}\itemsep=-1pt

\bibitem{Fernando2009_MMAC}
A.~Bargteil X. Martin J. Macey A.~Collado F.~Torre, J.~Hodgins and P. Beltran.
\newblock Guide to the carnegie mellon university multimodal activity
  (cmu-mmac) database.
\newblock In {\em Tech. report CMU-RI-TR-08-22, Robotics Institute, Carnegie
  Mellon University}, April 2008.

\bibitem{Sabour2017_CapsNet}
S. Sabour, N. Frosst, and G.~E. Hinton.
\newblock Dynamic routing between capsules.
\newblock In {\em Advances in Neural Information Processing Systems 30: Annual
  Conf. on NIPS, 4-9 Dec. 2017,}, pages 3859--3869, 2017.

\bibitem{Shi2017_convlstm}
X. Shi, Z. Gao, L. Lausen, H. Wang, D. Yeung, W. Wong, and W. Woo.
\newblock Deep learning for precipitation nowcasting: {A} benchmark and {A} new
  model.
\newblock In {\em Advances in Neural Information Processing Systems 30: Annual
  Conf. on NIPS}, pages 5622--5632, 2017.

\bibitem{VGG16}
K. Simonyan and A. Zisserman.
\newblock Very deep convolutional networks for large-scale image recognition.
\newblock {\em CoRR}, abs/1409.1556, 2014.

\bibitem{Soran15}
Shapiro~L. Soran~B., Farhadi~A.
\newblock Action recognition in the presence of one egocentric and multiple
  static cameras.
\newblock In {\em ACCV 2014. Lecture Notes in Computer Science,}, pages 17--24,
  2014.

\bibitem{Spriggs2009Temporal}
E.~H. Spriggs, F.~De~La Torre, and M. Hebert.
\newblock Temporal segmentation and activity classification from first-person
  sensing.
\newblock In {\em Computer Vision and Pattern Recogn. Workshops. IEEE Computer
  Society Conf. on}, pages 17--24, 2009.

\end{thebibliography}
}

\end{document}